\pdfoutput=1

\documentclass[11pt]{article}

\usepackage[]{ACL2023}

\usepackage{times}
\usepackage{latexsym}

\usepackage[T1]{fontenc}

\usepackage[utf8]{inputenc}

\usepackage{microtype}

\usepackage{inconsolata}

\usepackage{graphicx}
\usepackage{cleveref}
\usepackage{multirow}
\def\hideXXX#1{}
\def\XXX#1{{\textcolor{red}{XXX #1}}}  
\def\XXX#1{\hideXXX{#1}}

\usepackage[normalem]{ulem} 

\def\XXXifaccepted#1{{\textcolor{green}{Add if accepted: #1}}}
\def\XXXifacceptedWMT#1{}
\def\XXXifaccepted#1{} 

\def\hidePV#1{}

\def\footurl#1{\footnote{\url{#1}}}

\def\inparcite#1{\citealp{#1}} 
\def\furl#1{\footnote{\url{#1}}}

\title{Turning Whisper into Real-Time Transcription System 
}
\author{Dominik Macháček$^1$ \and Raj Dabre$^2$ \and Ondřej Bojar$^1$
  \\ \\
  Charles University, Faculty of Mathematics and Physics, \\ 
  Institute of Formal and Applied Linguistics$^{1}$ \\
   \\
  National Institute of Information and Communications Technology, Kyoto, Japan$^2$\\
  $^{1}$\texttt{\{machacek,bojar\}@ufal.mff.cuni.cz}, $^2$\texttt{raj.dabre@nict.go.jp}}

\begin{document}
\maketitle
\begin{abstract}
Whisper is one of the recent state-of-the-art multilingual speech recognition and translation models, however, it is not designed for real time transcription. In this paper, we build on top of Whisper and create \textbf{Whisper-Streaming}, an implementation of real-time speech transcription and translation of Whisper-like models. Whisper-Streaming uses local agreement policy with self-adaptive latency to enable streaming transcription. 
We show that Whisper-Streaming achieves high quality and 3.3 seconds latency on unsegmented long-form speech transcription test set, and we demonstrate its robustness and practical usability as a component in live transcription service at a multilingual conference.


\end{abstract}

\section{Introduction}

Whisper \cite{Whisper-paper} is a recent state-of-the-art system for automatic speech recognition (ASR) for 97 languages and for translation from 96 languages into English. Whisper models are publicly available under the MIT license. However, the current public implementations of Whisper inference usually allow only offline processing of audio documents that are completely available at the time of processing, without any processing time constraints.

Real-time streaming mode is useful in certain situations, e.g.\ for live
captioning. It means that the source speech audio has to be processed at the
time when it is being recorded. The transcripts or translations have to be
delivered with a short additive latency, e.g.\ in 2 seconds. There are some implementations of Whisper for streaming, but their approach is rather naive, they e.g.\ first record a 30-second audio segment, and then process it. The latency of these methods is large, and the quality on the segment boundaries is low because a simple content unaware segmentation can split a word in the middle.

In this work, we implement, evaluate and demonstrate Whisper in simultaneous
streaming mode using the simple but effective LocalAgreement
\cite{liu20sinterspeech} algorithm. LocalAgreement is one particular
streaming policy that can be used to convert any full-sequence to
full-sequence model to operate in simultaneous streaming mode. It was used
by the winning system CUNI-KIT at IWSLT 2022 simultaneous speech translation
shared task \cite{polak-etal-2022-cuni}. We call our implementation
\textbf{Whisper-Streaming}, although it is applicable to any model with API
similar to Whisper. According to our evaluation, it achieves 3.3 seconds
latency on average for English ASR on the European Parliament speech test
set ESIC \cite{machacek21_interspeech}, when running on NVIDIA A40 GPU, a fast hardware processing unit. We test it also on German and Czech ASR and present the  results and suggestions for the optimal parameters.

The contribution of this work is implementation, evaluation and demonstration of Whisper-Streaming. Given that Whisper-Streaming can be quickly and easily packaged into a product, we want to ensure that the most recent scientific results, such as the algorithm for simultaneous mode, can be accessible to and be used by industrial researchers and engineers. 
Furthermore, we want to reliably evaluate the performance of our implementation and share the results with the research community, to further drive research and development of real-time transcription solutions which have real-life use cases. We expect that our results can be used as strong baselines for future comparison.

We make Whisper-Streaming publicly available\footnote{\url{https://github.com/ufal/whisper_streaming}} along with a demonstration video.\footnote{\url{https://vimeo.com/840442741}}


\begin{figure*}[th!]
    \centering
    \includegraphics[width=0.95\textwidth]{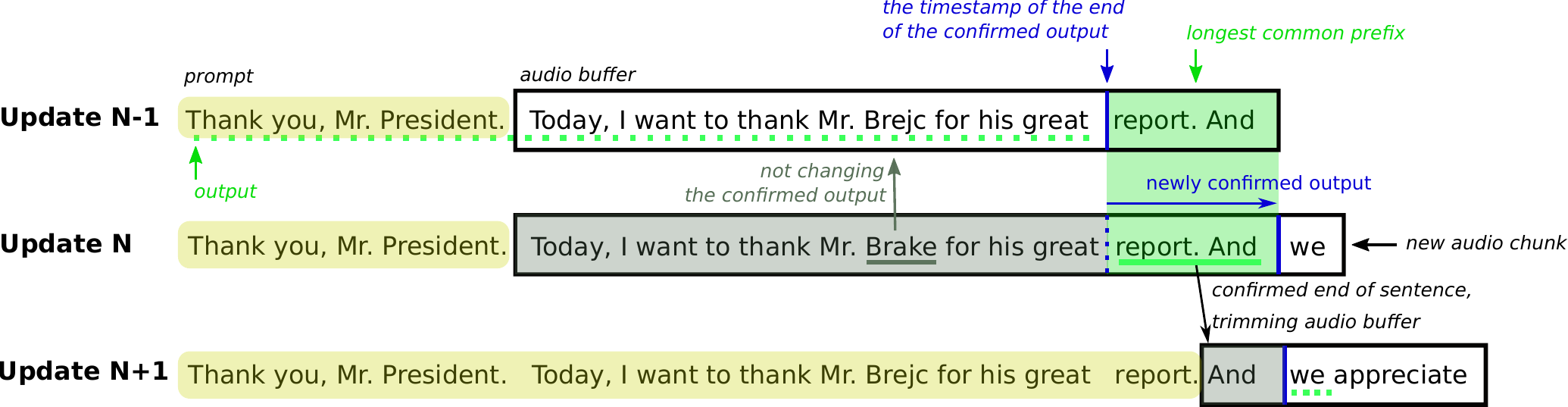}
    \caption{Illustration of processing three consecutive updates. The yellow highlighted text is a ``prompt'', the previous context to follow. The black-bordered rectangle is an audio buffer, and the text inside is Whisper's transcript generated from that sound segment. The blue vertical line is a timestamp that splits the buffer to two parts, the left being previously confirmed, and the right one is unconfirmed. The LocalAgreement-2 policy, or searching the longest common prefix, is applied on the unconfirmed (right) part in two subsequent updates. The longest common prefix is highlighted in green and the green underline highlights the \textit{newly} confirmed output, whereas the green dashed underline indicates previously and subsequently confirmed output. The gray underline demonstrates an update in the confirmed part that is disregarded.
    \XXX{if updating the picture, change the label of "end of the confirmed output" to "the timestamp of the end..."}
    }
    \label{fig:updates}
\end{figure*}

\section{Background}

In this section, we describe the background for the back-end components of our work.

\paragraph{Whisper} \cite{Whisper-paper} is a Transformer model for
speech-to-text transcription and translation trained on a massive amount of
multilingual data. We use
``large-v2''\footnote{\url{https://huggingface.co/openai/whisper-large-v2}}
model because it achieves the highest quality of all Whisper model size options. 
Since the original release of the whisper backend is rather slow, we use the \texttt{faster-whisper}\furl{https://github.com/guillaumekln/faster-whisper} reimplementation of Whisper inference using CTranslate2, a fast inference engine for Transformer models. It is approximately four times faster than the standard implementation (as reported by the authors). We use it with 16-bit float precision.

Although we primarily use Whisper, the underlying model in our implementation can be easily replaced by any other speech-to-text transcription or translation model (e.g.\ MMS, \inparcite{pratap2023mms}) if it produces word-level timestamps and punctuation.



\paragraph{Streaming} 


Let us assume a model $M$ that processes a source sequence $c_1,\cdots,c_n$
into a target sequence $t_1,\cdots,t_m$, given a previous target 
$s$ that can be used for for inter-sentence coherence. Streaming involves receiving the source sequence consecutively, one chunk at a time, and producing the target simultaneously. A \textit{streaming policy} $P$ predicts a target segment $t_T$ at time $T$ as $t_T := P_M(c_{i<T}|s,t_{j<T})$. It operates the model $M$ on available source chunks $c_{i<T}$, previous sequence target $s$, and previous target segments $t_{j<T}$. The policy is triggered every time a new source segment is available. An empty target segment can be emitted, e.g.\ when waiting for context. The policy aims to minimize latency and maximize target quality. 

Streaming was originally proposed for simultaneous translation \cite{ma-etal-2019-stacl}, but it is applicable for any sequence-to-sequence task including ASR. \citet{dong-etal-2022-learning} give a summary of streaming speech translation.

\paragraph{LocalAgreement} 
\cite{liu20sinterspeech} is a streaming policy that outputs the longest common prefix of the model on $n$ consecutive source chunks, or an empty segment when less than $n$ chunks are available. Based on the IWSLT 2022 shared task on simultaneous translation, the CUNI-KIT system compared LocalAgreement to other policies (hold-$n$ and wait-$k$) with different chunk sizes. They found that LocalAgreement with $n=2$ was the most effective policy. Therefore, we use LocalAgreement-2 for identifying stabilized target segments.

\def\param#1{#1}
\def\MinChunkSize{\param{MinChunkSize}}

\section{Whisper-Streaming}
We describe the core components and inner workings of Whisper-Streaming. It consists of the update loop, audio buffer, skipping the confirmed output in audio buffer, trimming the buffer, joining for inter-sentence context, and optional voice activity detection.



\paragraph{Update Loop}
The main part of Whisper-Streaming is a program that utilizes a loop to receive source audio chunks and trigger streaming policy updates. The parameter \MinChunkSize{} controls the latency and quality, and determines the minimal duration processed per iteration. If the update computation exceeds \MinChunkSize{}, the next update is performed immediately on the accumulated audio input. This parameter impacts both latency and quality.


\paragraph{Audio buffer}


Whisper is trained to handle sequences that are up to 30 seconds long and contain one full sentence. It provides punctuation and word-level timestamps.\footnote{When using ``faster-whisper'' or aannother implementation that supports it.} The process is illustrated in \Cref{fig:updates}. Each update involves storing incoming audio at the top of the audio buffer and processing the entire buffer with Whisper. We keep an invariant that the buffer always starts with a new sentence, to maintain the high quality of Whisper. LocalAgreement-2 is applied to the current and previous Whisper output. The timestamp of the last word in the ``confirmed output'' is saved. In subsequent updates, we always reprocess Whisper from the beginning of the buffer, including the portion preceding the last ``confirmed output'' timestamp (indicated by the gray background in \Cref{fig:updates}). Changes to the transcription in the confirmed portion are disregarded, as they are often insignificant in terms of meaning alteration.

\paragraph{Skipping the confirmed part}

When determining the position of transcribed words relative to the last confirmed word from the previous update, we account for the potential inaccuracies and updates in Whisper timestamps due to new audio chunks. If a word's timestamp falls within a 1-second interval from the last confirmed word, we compare its preceding $n$-grams (where $n$ ranges from 1 to 5) with the suffix in the last confirmed output. If they match, we skip those words. However, this rule can be further enhanced in future work by incorporating measures such as setting and fine-tuning a character edit distance threshold, trimming punctuation and casing from the $n$-grams, etc.

\paragraph{Trimming the audio buffer}


To avoid inacceptably long spikes in latenc, the audio buffer is limited to around 30 seconds. When the confirmed output includes a sentence-ending punctuation mark followed by a word starting a new sentence, the buffer is trimmed at the punctuation mark's timestamp. A language specific sentence segmentation tool (e.g.\ \inparcite{koehn-etal-2007-moses}) is used for this purpose, ensuring that the buffer always contains a single sentence. Despite this, if the buffer length exceeds 30 seconds, we retain the last confirmed segment marked by Whisper.


\paragraph{Joining for inter-sentence context}


The Whisper transcribe function utilizes a ``prompt'' parameter to maintain consistency within a document (consistent style, terminology, and inter-sentence references). We extract the last 200 words from the confirmed output of previous audio buffers as the ``prompt'' parameter, as shown in \Cref{fig:updates} (yellow backgrounded text).

\paragraph{Voice activity detection}
There is a parameter to activate or deactivate Whisper's default voice activity detection (VAD) filter, impacting both quality and latency.

\section{Benchmarking Settings}
We describe the dataset for evaluation, metrics, settings and hardware we used to evaluate our model.

\paragraph{Evaluation Data} 

For latency and quality analysis, we utilize the dev set of the manually transcribed ESIC corpus \cite{machacek21_interspeech} for English, German, and Czech ASR containing 179 documents. This corpus contains 5 hours of original English speeches from the European Parliament, including simultaneous interpreting into German and Czech. It provides audio tracks with manual transcripts and word-level timestamps. 


\paragraph{WER}
We use word error rate (WER) after removing punctuation and casing as the standard measure of ASR quality. 

\paragraph{Latency}


In our latency analysis, we implement our own method wherein we use the timestamps provided in the ESIC corpus to align the gold transcripts to the ASR output using edit distance.\furl{https://pypi.org/project/edlib/} This allows us to determine the edit operations for each gold word. We calculate the ASR latency by measuring the time difference between when the ASR emitted a word and when the corresponding gold word was spoken, excluding words deleted by the ASR. We compute the average latency within each document and, when comparing different setups across multiple documents, we report the average latency along with standard deviation.

\paragraph{Hardware}
For benchmarking, we use NVIDIA A40 GPUs. 
We run Whisper on a computer in a cluster that is used by other processes at the same time, which may allocate the same resources and influence the latency. Since it is not always possible to have a dedicated server for a given service, this makes our evaluation very realistic. Since there will be variations in the latency metrics, we report mean and standard deviations. 

\def\AForty{A40}
\def\LForty{L40}
\def\VADno{off}
\def\VADyes{on}
\def\pms{$\pm$}
\begin{table}[t]
    \centering
    \begin{tabular}{ll|rr}
  GPU & VAD &\% WER &  latency [s] \\
  \hline
\bf \AForty{} & \bf \VADno{} & \bf 5.8\pms{}0.9 & \bf 2.85\pms{}0.45 \\
\AForty{} & \VADyes{} & 5.2\pms{}0.9 & 3.12\pms{}0.36 \\
\LForty{} & \VADno{} & 5.1\pms{}1.0 & 3.58\pms{}0.62 \\
\LForty{} & \VADyes{} & 5.0\pms{}0.6 & 3.96\pms{}0.81 \\
    \end{tabular}
    \caption{Average (\pms{}stddev) WER and latency of English ASR of 10 repeated runs of ESIC dev.20080925.013\_007 document, with \MinChunkSize{} 0.1 seconds, using or not using VAD filter, on two GPU types. Bold is the setup that we later use.} 
    \label{tab:repro}
\end{table}

\paragraph{Ensuring Reproducibility} We simulate real-time processing of long-form transcription and record the  times when Whisper emitted the outputs.
We run the simulation on computers in a cluster that are not entirely under our control. For our simulation process, we block one GPU and a sufficient number of CPUs and RAM capacity. However, it can happen that other processes run at the same time, making a CPU and RAM load that is unpredictably slowing down our simulation. If the \MinChunkSize{} is smaller than time for processing an update, then two runs of the same simulation have different segmentation to chunks, leading to different WER and latency. 

Therefore, we run simulation of the same setup of one document 10 times, to measure the standard deviation of the latency and quality. The setup is English transcription of the ESIC dev.20080925.013\_007 document that is 3 minutes 36 seconds long, on NVIDIA A40 or L40 GPU with 48GB GPU RAM, 8 blocked CPU cores and 200GB of CPU RAM, with or without VAD filter, with \MinChunkSize{} 0.1 seconds.

The results are in 
\Cref{tab:repro}. We observe small, negligible standard deviation in WER, below or near 1\%. The standard deviation in the average latency is much larger, from 0.36 to 0.81 seconds depending on the setup. 
We conclude that we must be aware of the standard deviation of latency due to uncontrollable computation conditions. 

\section{Results}

We evaluated Whisper-Streaming with various setups for English, German and Czech ASR. We first show the impact of outliers and voice activity detection (VAD) to determine optimal settings, and then present our main results with these settings.

\paragraph{Outliers} 
After processing many setups, we observed extraordinarily high WER on English ASR of a document dev2.20101213.015\_018\_EN\_Gallagher. We realized it is due to noise in the ESIC data set. The first half of the mentioned document is in Irish, and not English as intended. Only the English part is transcribed in gold, but Whisper transcribed both, leading to more precise transcription than the reference.
Except of the Gallagher document, all the reported setups achieved WER between 0 and 52\%, and average latency between 0 and 16.1 seconds. 

\begin{figure}
    \centering
    \includegraphics[width=\columnwidth]{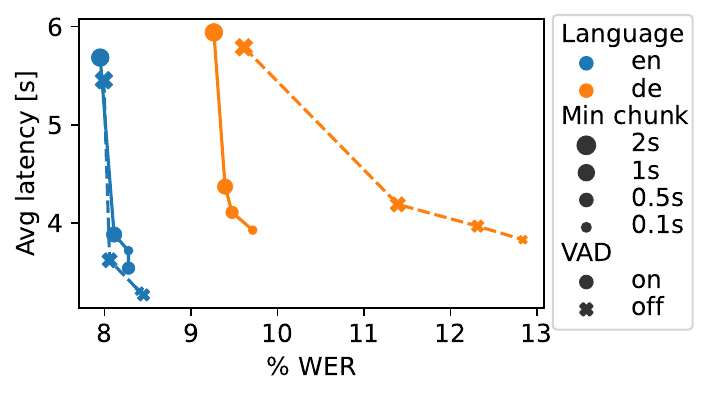}
    \caption{Impact of VAD filter on latency and quality. The striking difference in VAD activated or deactivated for English vs German is due to German being the speech of an interpreter.}
    \label{fig:vad}
\end{figure}

\paragraph{VAD} 
We studied the effect of VAD (voice activity detection) filter that is integrated within Whisper backend. The results are in \Cref{tab:vad} and \Cref{fig:vad}. We realized that in ESIC corpus, it is advisable to deactivate the VAD filter for the English original speech, because it is very fluent, not interleaved with silence and has no non-voice sounds. Without VAD, the quality remains nearly the same (difference within 0.2\% WER), and the average latency was substantially lower, between 0.23 to 0.41 seconds.

For the processing of simultaneous interpreting, we recommend activating the VAD filter. The speech of a simultaneous interpreter contains many pauses, especially when waiting for context. With VAD, the latency was only 0.1 seconds larger, because VAD often filters out silence, which reduces the processing load. The quality with VAD was substantially higher, by 2 to 3 \% WER with shorter \MinChunkSize{} on German. With large chunk sizes, the quality is nearly the same (0.3 \% WER difference with 2 seconds \MinChunkSize) because a large chunk size causes the model to have large context and thus a low chance for risking uncertain output. Therefore,  we activated VAD for German and Czech simultaneous interpreting, and we deactivated it for English original speech.

For a real-life setup, we recommend starting Whisper-Streaming shortly before the speech actually starts, so that the first words are not missed, along with turning the VAD filter on so that the silence and non-voice sounds do not cause Whisper to make mistakes. If reducing the latency is important, an adaptive protocol for setting VAD on and off can be implemented. 


\def\LanDe{de}
\def\LanEn{en}
\def\LanCs{cs}
\def\VADyes{on}
\def\VADno{off}
\begin{table}[t]
\footnotesize{}
    \centering
    \setlength{\tabcolsep}{4pt}
    \resizebox{1.0\columnwidth}{!}{%
    \begin{tabular}{c@{~}c||rr|r||rr|r}
   &        & \multicolumn{3}{c||}{avg.\ \% WER} & \multicolumn{3}{c}{avg. latency [s]} \\
 & m.ch. & \VADno{} & \VADyes{} & diff & \VADno{} & \VADyes{} & diff \\
\hline
\multirow{4}{*}{\LanEn{}} 
      & 0.1s   & 8.4 &  8.3 & -0.1  &  \bf 3.30 & 3.72 & +0.41 \\
      & 0.5s   & 8.5 &  8.3 & -0.2  &  \bf 3.27 & 3.54 & +0.27 \\
      & 1.0s   & 8.1 &  8.1 & +0.1  &  \bf 3.62 & 3.88 & +0.26 \\
      & 2.0s   & 8.0 &  7.9 & -0.0  &  \bf 5.45 & 5.68 & +0.23 \\
\hline
\multirow{4}{*}{\LanDe{}} 
      & 0.1s   & 12.8 & \bf 9.7 & -3.1  &   3.83 & 3.93 & +0.10 \\
      & 0.5s   & 12.3 & \bf 9.5 & -2.8  &   3.97 & 4.11 & +0.14 \\
      & 1.0s   & 11.4 & \bf 9.4 & -2.0  &   4.19 & 4.37 & +0.18 \\
      & 2.0s   & 9.6 & \bf 9.3 & -0.3  &   5.79 & 5.94 & +0.15 \\
    \end{tabular}}
    \caption{Impact of VAD filter on WER and latency on ESIC dev on the streaming ASR with different minimum chunk size (m.ch., in seconds) of the English original speech (\LanEn{}) and German simultaneous interpreting (\LanDe{}). 
    We highlight the remarkable benefit in bold: the original speech without pauses is processed with lower latency (by 0.23 seconds or more) and comparable quality with VAD off. On the contrary, the VAD on achieves higher quality for interpreting with frequent pauses, with small difference in latency.
    }
    \label{tab:vad}
\end{table}

\paragraph{Performance}



\begin{figure}[t]
    \centering
    \includegraphics[width=\columnwidth]{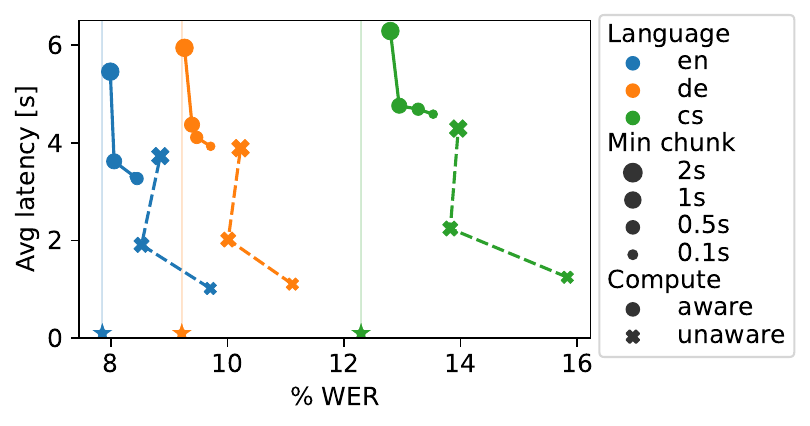}
    \caption{Latency and quality in computationally aware and unaware simulations (solid lines and dots vs dashed lines and crosses), together with offline WER (stars and light vertical lines). 
    VAD is deactivated for English, and activated for the other two.
    }
    \label{fig:bounds}
\end{figure}

\Cref{tab:unaware} and \Cref{fig:bounds} summarize the WER and average latency of Whisper-Streaming on ESIC validation set for the three language tracks. Overall, with 1 second \MinChunkSize{}, the average latency is 3.3 seconds for English, 4.4 seconds for German and 4.8 seconds for Czech, while the WER is by 2\% higher than in the offline mode for English and German, and by 6\% higher for Czech. 
Both WER and latency is the lowest on English, followed by German and Czech. This is related to the amount of language specific data used for training Whisper, as well as the morphological complexity of these languages. The latency increases with larger uncertainty because it requires more updates for an agreement. Moreover, the larger \MinChunkSize{}, the larger the latency, but higher the quality because the system has sufficient context. 

\paragraph{Offline mode WER}
We contrast the results with setups that serve as maximum performance estimates. One of them is offline mode in which processing of the whole audio document is done after recording, without any limitations on processing time. It is the default and most optimized setup for Whisper. The WER in offline mode and with VAD is lower than in streaming mode because the context size is not restricted. The model can use even the right (future) context that is unavailable or limited in streaming mode. Moreover, the internal segmentation of the long-form speech into processing chunks is optimized in the offline mode.

\paragraph{Computationally unaware latency}
Another contrastive setup is computationally unaware simulation. It uses an unrealistic assumption that computation for Whisper processing any audio segment is instant, so that the latency caused by computation is not included in the latency measurement. The measurement includes latency caused by uncertainty in the language. The gap between latency in computationally unaware and aware evaluation can be reduced by optimizing the hardware or inference algorithm. Computationally unaware latency can be reduced by improving the model or streaming policy.

We observe that the average computationally unaware latency is approximately twice the chunk size. This is expected because we use local agreement of two consecutive updates. However, the processing of English is actually faster, little less than twice the chunk size. We hypothesize that this could be caused by the anticipation ability of Whisper model. The second possible reason is the inaccuracy of the gold timestamps in ESIC. The timestamps were computed by automatic forced alignment, and thus they may be less accurate in non-standard situations such as overlapping and non-transcribed speech, e.g.\ hesitations and foreign language insertions.

\def\unaware{un.}
\def\aware{aw.}
\begin{table}[h]
    \centering
    \setlength{\tabcolsep}{4pt}
    \resizebox{\columnwidth}{!}{
    \begin{tabular}{cc|r|rr|rrr}
    &  \% WER       &       & \multicolumn{2}{c|}{\% WER} & \multicolumn{3}{c}{latency [s]} \\
lang. & offline & m.ch. & \unaware{} & \aware{} &  \unaware{} & \aware{} & diff \\
\hline
\multirow{3}{*}{\LanEn{}} & \multirow{3}{*}{7.9} 
      & 0.5s   & 9.7 & 8.5   &   1.02 & 3.27 & +2.25 \\
   &   & 1.0s   & 8.5 & 8.1   &   1.91 & 3.62 & +1.71 \\
   &   & 2.0s  & 8.8 & 8.0   &   3.73 & 5.45 & +1.73 \\
   \hline
\multirow{3}{*}{\LanDe{}} & \multirow{3}{*}{9.2} 
      & 0.5s   & 11.1 & 9.5   &   1.11 & 4.11 & +3.00 \\
   &   & 1.0s   & 10.0 & 9.4   &   2.02 & 4.37 & +2.35 \\
   &   & 2.0s   & 10.2 & 9.3   &   3.89 & 5.94 & +2.05 \\
   \hline
\multirow{3}{*}{\LanCs{}} & \multirow{3}{*}{12.3} 
      & 0.5s   & 15.8 & 13.3   &   1.25 & 4.69 & +3.44 \\
  &  & 1.0s   & 13.8 & 12.9   &   2.24 & 4.76 & +2.51 \\
    &  & 2.0s   & 14.0 & 12.8   &   4.29 & 6.29 & +2.00 \\
    \end{tabular}}
    \caption{WER and average latency of Whisper-Streaming on ESIC dev set in three language tracks using different \MinChunkSize{} (``m.ch.''). The realistic setup is computationally aware (``\aware{}''), put into contrast with 
    offline WER (``offline'') and with the computationally unaware simulation (``\unaware{}'').
	The data are the same as in \Cref{fig:bounds}.
	}
    \label{tab:unaware}
\end{table}













\section{System Demonstration}

\paragraph{Demonstration video} is available at \url{https://vimeo.com/840442741}. It is a screencast video of Whisper-Streaming real-time outputs that processes live ASR on one ESIC document in three parallel instances for English, German and Czech speech, the original and simultaneous interpreting. The video shows a contrast to gold transcripts with original timing, so that the latency can be observed. The video also contains color highlighting for ASR errors.

\paragraph{Integration with ELITR} To demonstrate practical usability, we integrate Whisper-Streaming with the ELITR (European Live Translator, \inparcite{bojar-etal-2020-elitr}) framework for complex distributed systems for multi-source and multi-target live speech transcription and translation \cite{bojar-etal-2021-elitr}. Within Whisper-Streaming, we implement and release a server that is connected as a worker to Mediator server \cite{franceschini-etal-2020-removing}. Mediator allows a client to request a service of a worker. The client is then allowed to further process the text outputs received by the worker, e.g.\ translate them with another worker and present them at the web view server that delivers real-time captions to event participants during a live multilingual event. 

\paragraph{Evaluation event}
We evaluated Whisper-Streaming as a component in an experimental live speech translation service at a multilingual conference. For this, we built a pipeline that used five parallel Whisper-Streaming workers, three of them for ASR only (English, Czech and Ukrainian), and two for speech translation (Czech-to-English and Ukrainian-to-English). There were three parallel language streams at the conference, Czech, English and Ukrainian. One of the languages was spoken at the main floor, and the others were provided by human simultaneous interpreting. 

A human operator (as in \inparcite{bojar-etal-2021-operating}) was controlling the technical setup and the outputs using the language knowledge and had an option to redirect the streams, if necessary.
The qualitative evaluation at the event showed that Whisper-Streaming is a robust and reliable part of the service, reaching acceptable latency and unexpectedly high quality on English, Czech and Ukrainian long-form speech.

\paragraph{Demonstration at AACL}
Our system demonstration at the IJCNLP-AACL 2023 conference will use the ELITR framework. We will either simulate speech source from a recording, or allow participants to speak into microphone in any of the 97 languages supported by Whisper, and observe the real-time outputs.

\section{Conclusion}

We implemented, evaluated and demonstrated Whisper-Streaming, a tool that effectively operates an offline ASR model Whisper, with 3.3 second average computationally aware latency on English ESIC corpus. 
We described and explained the implementation and its underlying components, including LocalAgreement algorithm for streaming. Lastly, we demonstrated the robustness and practical usability at a real-life multi-lingual conference.


\section{Limitations}







The data collected in ESIC corpus were created relatively long time ago. It raises concerns about potential leakage into Whisper training set, which could compromise our evaluation. 
Additionally, performance tests on more affordable hardware are pending, highlighting the need for further evaluation in terms of computational cost.

It is worth noting that the reported latency and quality metrics obtained from ESIC may not be fully generalizable to other languages or language variants due to the nature of the corpus.  

Furthermore, our focus is on demonstrating the online capabilities of Whisper rather than optimizing the algorithm or implementation. It is important to recognize that the actual latency experienced may fluctuate, and the reported average latency serves as an indicative measure without providing an upper bound. The streaming policy would need certain modifications to guarantee a maximum latency, at a possible loss in quality.


Lastly, we have not conducted comparison tests to other state-of-the-art systems, e.g.\ from IWSLT, because a common evaluation framework is pending, as well as X-to-English long-form speech test set.

\section*{Acknowledgements}

\XXX{}
This research was partially supported by the grants 19-26934X  (NEUREM3)  of  the  Czech  Science Foundation, 
and SVV project number 260~698.

\bibliography{anthology,custom}
\bibliographystyle{acl_natbib}

\end{document}